\documentclass[runningheads]{llncs}

\usepackage{times}
\usepackage{textcomp}
\usepackage{graphics} 
\usepackage{epsfig} 
\usepackage[svgnames]{xcolor}
\usepackage{amsfonts}
\usepackage{breqn}
\usepackage{algorithm}
\usepackage{algpseudocode}

\usepackage{setspace}
\usepackage{placeins}

\usepackage{subcaption}
\usepackage{booktabs}
\usepackage{multirow}
\usepackage{url}
\usepackage{gensymb}

\usepackage{textcomp}
\definecolor{comments}{rgb}{0.0,0.7,0.0}

\begin{document}
\title{Proactive Intention Recognition for Joint Human-Robot Search and Rescue Missions through  Monte-Carlo Planning in POMDP Environments}
\titlerunning{Proactive Intention Recognition for Joint Human-Robot Search and Rescue Missions}
%
\author{Dimitri Ognibene\inst{1}\orcidID{0000-0002-9454-680X} \and
Lorenzo Mirante\inst{1}
\and
Letizia Marchegiani\inst{2}\orcidID{000-0001-6782-6657}
}
\authorrunning{D. Ognibene, L. Mirante \& L. Marcegiani}
%
\institute{School of Computer Science and Electronic Engineering,
University of Essex, Colchester, UK
\email{dimitri.ognibene@essex.ac.uk}\\
\and
Department of Electronic Systems, Aalborg Univeristy, Aalborg, DK\\
\email{lm@es.aau.dk} 
}
\maketitle              
\begin{abstract}
Proactively perceiving others' intentions is a crucial skill to effectively interact in unstructured, dynamic and novel environments.
This work proposes a first step towards embedding this skill in support robots for  search and rescue missions. 
Predicting the responders' intentions, indeed, will enable exploration approaches which will identify and prioritise areas that are more relevant for the responder and, thus, for the task, leading to the development of safer, more robust and efficient joint exploration strategies.
More specifically, this paper presents an active intention recognition paradigm to perceive, even under sensory constraints, not only the target's position but also the first responder's movements, which can provide information on his/her intentions (e.g. reaching the position where he/she expects the target to be). This mechanism is implemented by employing an extension of Monte-Carlo-based planning techniques for partially observable environments, where the reward function is augmented with an entropy reduction bonus.
We test in simulation several configurations of reward augmentation, both information theoretic and not, as well as belief state approximations and obtain substantial improvements over the basic approach.
\end{abstract}
%
%
%

\section{INTRODUCTION}
Humans are endowed with sophisticated social interactions skills that provide invaluable advantages. These are realised by  implicit intention reading,  through the interpretation of partners' sensorimotor coupling with the environment, or explicit communication 
\cite{pezzulo2019body,ognibene2019implicit}. The former modality can be advantageous for the `acting' partners as it requires a lower cognitive load and allows focusing on the task at hand.  However, the perceiving partner must gather and integrate information on several important factors like the presence, identity and configuration of  any relevant affordances, as well as the pose of the partner and the trajectories of its effectors \cite{ognibene2013contextual}. In complex and unstructured environments these skills have to heavily rely on proactive perception capabilities, as not all these elements may  be immediately and simultaneously observable (\textit{e.g.} being in another room, or occluded, out of the field of view or simply unattended), and require multiple steps to be sensed. This is one of the main issues impeding human robot collaboration in such environments as dense sensorisation of the environment necessary for passive perception is not viable, and current active perception strategies are still limited to single step `myopic' proactive perception strategies \cite{Ognibene2013,Lee2015}.  
An important application domain for this skill is search-and-rescue missions where both risk for the  first responders and time spent in localising a survivor must be optimized.
Yet, fully automating  even the riskiest and earliest phases of these tasks is still hindered by limited autonomous exploration algorithms. Human  responders, indeed, rely not only on the flexibility of their perceptual skills, but also on non-verbal task knowledge gained by exploring multiple  environments, which is difficult to transfer to a robot. 

In this work we propose the use of active intention recognition (IR) paradigms to allow the robot to indirectly and naturally exploit first responders' knowledge and, thus, to improve its exploration performance.
The intuition is that the responders may be able to more accurately guess the position of the target and start moving to reach it. Their initial movements, together with some partial information on the map (\textit{e.g.} information on regions of interest), can be enough to estimate where the responders are directed and, thus, anticipate them to test the location and secure the path.
We carry out our analysis in simulation assuming the use of a drone as the search-and-rescue robot of reference.

\begin{figure}[t]
\centering
\includegraphics[width=\linewidth]{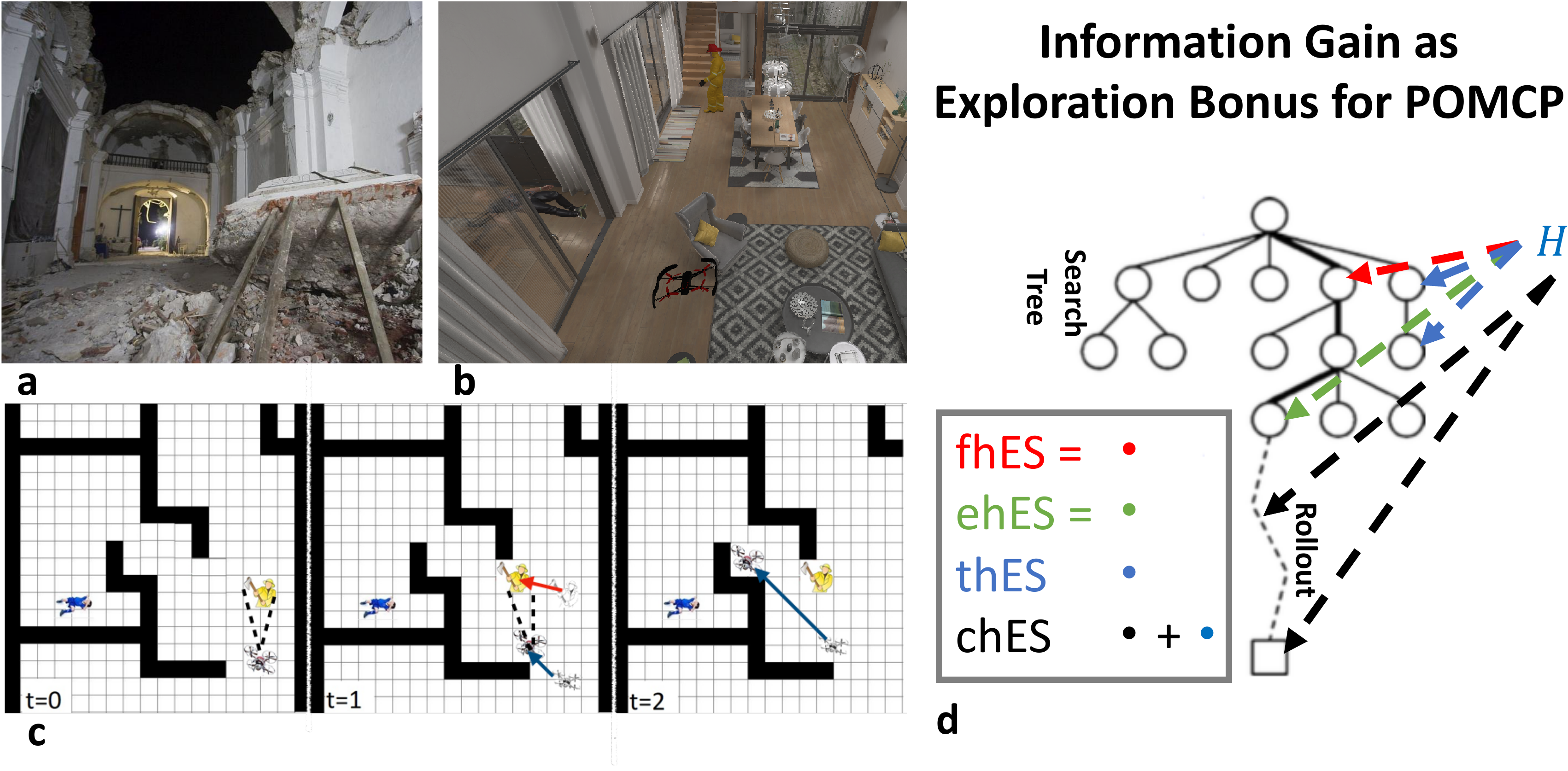}
\caption[Intention Recognition for Collaborative Search Task Description]{\footnotesize{\textbf{a)} A real disaster condition. \textbf{b)} A simulated environment view generated by AIRSIM \cite{airsim2017fsr}  for testing our algorithm. The figure shows the drone, the responder and the target (\textit{e.g.} a survivor in a disaster). The responder is supposed to be able to easily guess a good location where the target might be and slowly approach it. The drone enters the disaster zone from a different position, has initial knowledge of the entrance location of the responder and of a set of candidate areas where the target could be located. The drone, due to the complexity of the environment (\textit{e.g.} occlusions), can see the target only when it is close to it. It must employ a policy that can take advantage of the presence of the responder (whose original position is known) to find the target as fast as possible. \textbf{c)} A cartoon-like illustration of an example scenario. \textit{t=0)} the drone finds the first responder; \textit{t=1)} the drone observes the responder moving towards the room where the survivor is; \textit{t=2)} the drone anticipates the first responder and approaches the survivor room.} \textbf{d)} Information gain as exploration bonus for POMCP. The arrows represent where in the MCTS search three is computed the entropy used for  the exploration bonus.}
\label{fig:task-description}
\vskip-5ex
\end{figure}

\section{Background}
\subsubsection{Search and rescue} Robot disaster responders are an important research target \cite{Krujiff2012RescueRobots}. They aim to quickly locate and extract survivors. This has driven the study of the non-verbal interaction strategies adopted by human responders \cite{Bacim2012} and the development of collaborative navigation algorithms for groups of robots \cite{Beck2016Online}. 
In  this line of research we try to bridge these problems by enabling robots to more naturally understand and account for responders' intentions while finding computationally affordable solutions. 
\subsubsection{Planning in partial observable conditions}
Planning tasks are traditionally represented and addressed relying on \emph{Markov Decision Processes} (MDP), as being a framework which allows for a neat problem formulation, while also accounting for the presence of uncertainty and control noise. Most commonly, in MDPs, the task is defined as finding a \textit{policy }that maximizes the \textit{expected cumulative discounted  reward }given a representation of the environment in terms of (i) a set of states and (ii) actions, (iii) a transition function describing the probability of reaching a certain state once an action is performed in a state, and a (iv) reward function describing the reward expected once an action
is performed in a state.
More realistic setups require to model noisy or missing sensory information.  These problems are described extending MDPs to \emph{Partially Observable MDP} (POMDP) \cite{KaelblingLittmanEtAl1998}, by providing an (v) observation function which describes the probability of obtaining a certain sensory input when the environment and the agent are in certain configurations. POMDP are particularly suitable to describe search-and-rescue environments, where the positions of the targets and the one of the responder are not necessarily known and they are detectable only when the robot is in their proximity. In this work, to approach these issues, we extend POMCPs, a Monte Carlo Tree Search (MCTS) algorithm for POMDP, which has been proposed in \cite{Silver2010}.
\subsubsection{Intention Recognition (IR)}
IR supports natural, proactive collaboration between robots. Several algorithms have been proposed in the literature: some based on the idea of using pre-baked libraries of actions which suit a specific condition; others relying on the idea of inverse planning and the possibility of predicting the intentions of other agents, given a model of the environment which can be used to evaluate the effectiveness of a sequence of actions to achieve any possible goals \cite{Baker2017RationalQuantitative,Ramirez2009}. This kind of approach leads to a more precise and flexible recognition process, but has a higher computational cost. Other approaches, particularly suitable for robotics applications, instead, employ the idea of IR as an inverse control task, where multiple parametrizable controllers are compared to explain others' behaviours \cite{Ognibene2013,Demiris2007}.
\subsubsection{Active Perception and Active Vision}
Active Perception is  the problem of actively controlling sensors to improve the speed and accuracy in the estimation of behaviourally relevant variables \cite{Ognibene2015,SpragueBallard2004}. Its applications are numerous, spanning from inspection and localization, to object recognition and autonomous driving, and not only limited to robotics. Indeed, relevant efforts can be also be found in the neuroscience and cognitive science communities \cite{Friston2015}. Different approaches have been proposed in the literature: \textit{e.g.} methods based on neural models often trained by Reinforcement Learning (RL)  \cite{Ognibene2015,PalettaFritzEtAl2005,SpragueBallard2004}, evolutionary algorithms \cite{Croon2008a}, and information theoretic approaches \cite{DenzlerBrown2002,Ognibene2013}.
\subsubsection{Active Intention Recognition (AIR)}
Intention Recognition (IR) is a particularly challenging problem, as it needs to take into account the interactions among multiple elements (\textit{e.g.} agents, affordances and effectors) which may not be known or  observed simultaneously (\textit{e.g.} while watching only the mouse running, we may not see the cheese, or not notice that a cat is running after the mouse).
In unstructured and not instrumented environments, such as construction or disaster sites, or in general, when only few sensors with a limited receptive field are available,  IR can be impossible to achieve if those sensors are not efficiently directed, so that they can perceive and track other agents, their effectors and the affordances they are interacting with. 
Active intention recognition (AIR) has been recently introduced  as the problem of recognising the intentions of  other  agents, either humans or robots, when the  sensing process can actively be controlled to deal with missing observations \cite{Ognibene2013,ognibene2013contextual} or limited  bandwith \cite{Lee2015}.  
AIR is formally defined as selecting the policy of sensor control which minimizes the final expected entropy on the state of the observed agent.
Ognibene \& Demiris in \cite{Ognibene2013} presented an implementation of the concept on a humanoid robot, where IR is approached as an inverse control problem. The formulation  used a mixture of Kalman filters to represent the robot's possible movements, assuming they were generated by one of multiple parametrizable linear controllers,  and using a single-step myopic strategy based on a novel approximation of the  information gain on the mixture selector variable. In \cite{Lee2015} a similar method was employed to simultaneously recognize the  complex activities of multiple actors, represented using Probabilistic Context Free Grammars.

In this work we extend \cite{Ognibene2013}  by: a) closing the loop between social perception and interaction through the integration of AIR and action execution, aimed at the shared goal of finding the target (survivor); b) relaxing the myopic constraint and considering multi-step policies.
%
\section{Methods}
\subsubsection{Environment}
The problem is described according to the POMDP formalism through transition, observation  and reward functions. A termination function is also used, as adopted in other POMDP problems with Monte Carlo (MC) methods.
%

\textit{Transition function}
The environment is static apart from the responder's movements. The transition function is, thus, defined by using a simple, deterministic, model of the drone's  movements and a probabilistic model of the movements of the responder when searching and approaching the target. 
The state is composed of a 6 dimensional tuple, containing the position of the drone, the position of the responder and the one of the target. The responder either stays still with probability equal to $p_{still}$ (to simulate a slower speed compared to the drone) either  moves one step in the direction of the survivor location with $0.95(1-p_{still})$ or in random direction. 

\textit{Observation model}
The observation model  always provides the position of the drone without noise. The positions of the target and of the responder, instead, are available when they are in the same position of the drone.
%
%
%

\textit{Reward function}
The main objective for the system is  to find  the goal, which is the  same objective of the responder. A reward equal to 1 is given when the drone finds the goal. 
AIR is auxiliary to the main collaborative objective and, thus, rewarding social interaction directly is not part of the problem description. Yet, as explained in the algorithm subsection, several forms of intrinsic reward  \cite{Mirolli2014} are employed to address several algorithmic limits in exploring large environments and in exploiting the information provided by the responder's movements.

\textit{End function}
End function stops both the real world simulation and the simulation inside the MC sampling algorithm when the drone is near the target.
\subsubsection{Planner}
The controller derives from the POMCP algorithm \cite{Silver2010}. It integrates the planning  with the state estimation processes by using the  Monte Carlo simulation sampling to generate samples for the particle filter. POMCP has improved state-of-the-art performance in several tasks, and it has been already applied to social robotics in several instances \cite{Goldhoorn2014}. POMCP uses a black-box generative model of the environment to estimate the action-observation sequences' values through repeated sampling. Each sampling iteration starts from a root node associated with the current sequence of observations $h$ integrated in the belief state $b(h)$ with the initial `prior' belief $b_0$. Then, POMCP    builds a limited portion of the tree of the future actions and observations through the following steps:  1) an initial state is sampled from the current belief state $b(h)$; 2) an action $a$ is selected according to an action-selection strategy that balances exploration and exploitation; 3) an observation $o$ is obtained defining a new history $h'=[h,a,o]$,  the corresponding node is evaluated recursively if it is already in the search tree (going back to step 2), or is inserted into the search tree after getting a first estimate of its value by continuing the simulation using a predefined `rollout' action selection policy until a certain depth $d$; and 4) update the statistics of the tree nodes by back-propagating the simulation results up to the root.
%
%
%
\subsubsection{Modifications of POMCP}
\textbf{\textit{I. Exploration Reward}}
The default exploration strategy (\textbf{dES}) of POMCP algorithm \cite{Silver2010} relies on the Upper Confidence Bound (UCB) exploration bonus \cite{Auer2002} in action selection. In  environments where most of the rewarding/punishing states are deeper in the search tree than maximum search depth applying MCTS planning based only on UCB for exploration can be problematic. 
In these contexts, MCTS algorithms may fail to estimate the value of observing informative cues and even prefer useless actions, as staying still at the entrance, because they never simulate long enough to reach the reward and, thus, fail to plan and observe such cues.
In this specific task, the informative cues are the responders' movements, which inform on their intentions and goals \cite{Ognibene2013,Ramirez2009,Baker2017RationalQuantitative} but can be noisy and expensive to acquire.
 Trying to identify those intentions requires finding and following the responder, which can be quite inefficient in several cases, \textit{e.g.} when the drone is certain of the goal position  or near  to a candidate  position, and approaching it directly would be faster than relying on the responder. Thus, while in previous works involving AIR \cite{Ognibene2013,Lee2015}  the expected information gain on the others' intentions was driving the robot behaviour, here \textit{dealing with the trade-off between acquiring information about the responder and exploring autonomously} becomes an important part of the task.
To address these issues, five other specific exploration strategies are tested that use an `intrinsic' reward \cite{Mirolli2014,Bellemare2016}  to facilitate not only visiting new useful positions, but also the extraction of information from the  human rescuer's presence: \textbf{1.} the \textbf{r}esponde\textbf{r} observation reward (\textbf{rrES}), where a reward of $0.1$ is added when the robot observes the responder; \textbf{2.} \textbf{c}omplete entropy (\textbf{chES}), at each step (both tree search and rollout) an intrinsic reward is added equal to $-0.2 \cdot H$ where $H$ is the entropy (see fig. \ref{fig:task-description}.d); \textbf{3.} search \textbf{t}ree entropy (\textbf{thES}), like chES, but entropy is added only inside the search tree but not in the rollout avoiding the update of the filter; \textbf{4.} \textbf{e}nd search tree entropy (\textbf{ehES}), like shES, but entropy bonus is added only at the last step of the tree search, before the rollout phase. This makes the amount of bonus added independent from the number of search steps, factor that biases the other approaches; \textbf{5.} \textbf{f}irst step search tree entropy (\textbf{fhES}), where the bonus is added only after the first step of search in the tree. The effect of computing entropy only on the goal location (\textbf{gH}) like in \cite{Ognibene2013}, or on all the complete belief state (\textbf{bH}) is also tested for all the entropy based  exploration strategies. This last variation exploits the structured form of the belief space containing the drone, the rescuer and the goal's location. This is motivated by the low variability of goals compared with the trajectories followed by the human responder to achieve them. Such variability induces a high entropy that may affect the planning algorithm. Entropy is directly computed from the belief state. 
Inside the search tree, filter and entropy are cached  to speed up the algorithm.

We must note that using information gain instead of the reward for helper observation may be more computationally expensive but it avoids suboptimal behaviours, such as following the slower responder after the target location is known. Furthermore, the complex computations for information gain used in \cite{Ognibene2013} are not necessary anymore as an expectation of the information gain will be provided by the backup phase of MCTS.
%

\textbf{\textit{II. Rollout Policy}}
Together with a  basic random rollout policy (\textbf{rRS}) that simply randomly samples actions during the rollout phase,  five  rollout stategies are developed which exploit additional task knowledge to reach a possible survivor position during the rollout. They differ in the way this position is selected: 
1) \textbf{s}ample \textbf{t}arget \textbf{r}ollout \textbf{s}trategy (\textbf{stRS}), which selects the survivor position of the current Monte-Carlo sample state \footnote{POMCP samples a state from the initial belief state at the beginning of each iteration} as the target position for the robot movement during the rollout; 
 2) \textbf{d}eterministic \textbf{n}ear {t}arget (\textbf{dnRS}), which selects the nearest survivor position between the not visited ones as the objective of the robot's movement in the rollout; 
 3) \textbf{s}tochastic \textbf{n}ear target (\textbf{snRS}), which selects one of the not visited targets as the target of the robot's movement with a probability inversely proportional to the distance from the drone; 
 4) \textbf{d}eterministic most \textbf{p}robable target (\textbf{dpRS}), which selects the not visited target which has the highest probability of containing the goal. This is based on the filter estimation computed during the simulation phase; 
 5) \textbf{s}tochastic most \textbf{p}robable target (\textbf{spRS}), which acts similarly to dpRS, but samples stochastically a not visited survivor location according to its probability of being the goal location.  For all these five strategies, a new target is selected when the previous one is reached without finding the goal. 
Two policies for the actual \textit{action selection} are adopted  in combination with the above strategies for target selection: I. best rollout action (\textbf{bRA}), which deterministically select the best action to reach the selected target, and II. stochastic rollout action (\textbf{sRA}), which  selects an action stochastically with a probability inversely proportional to the distance between the reached state and the  selected target. To implement efficiently these modifications, the best actions to reach each target are  precomputed. 

\textbf{\textit{III. State estimation and reconstruction}}
While the algorithm employed is inspired by the POMCP,  it does not use a particle filter, instead it uses a \textbf{c}omplete discrete Bayesian \textbf{f}ilter \textbf{(cF)}. 
This helps deal with the limited amount of information the robot has access to during the task and which may strongly affect the particle filter performance.
The cF allows also for effective entropy computation as well, but still presents some limitations (for more details, see \cite{lauri2016planning,Boers2010}). 
To decrease its computational cost we implement a truncated filter (\textbf{aF}), which, after each update, only maintains the $N_s=20$ states with the highest probability and zeroes the others\footnote{Entropy is computed before this zeroing process}. 
Given that the movements are localized, \textit{i.e.} only adjacent cells can be accessed from another cell, the filter update cost is limited. However, this approximation may lead to observation conditions 
which are wrongly estimated as impossible, \textit{i.e.} to have zero probability (similarly to the particles impoverishment in particle filters). This happens when the robot either  \textit{i.} focuses all the probability on solely one target and finds it empty, or \textit{ii.} observes the first responder in an unexpected position. In these conditions, a new  belief state is re-sampled as follows:

I. when all the probability is focused on one false target, the belief is regenerated by simulating the first responder's behaviour  starting from its last observation at time $t_o$ and moving towards each of the non visited target locations $l_g$. The same probability $p_g(t_o)$, which was assigned to the corresponding targets at time $t_o$, is then assigned to each of these generated states $x_g$ . If the first responder was never observed, then the behaviour is simulated  from each possible starting condition $l_s$. 

II. When the first responder is observed in a non-expected position $l^*$, a new element of the state $x_g$ is generated for each of the non explored targets $l_g$. In this case the probability associated with that state is computed according to how efficient it is to visit $l^*$ for the first responder from its previous location $l_{old}$ before reaching the target location $l_g$, versus travelling directly towards $l_g$ \cite{Ramirez2011,Ramirez2009}. Thus $p(x_g)=p_g(t_o)\cdot c(l_{old},l_g)/\left[c(l_{old},l^*)+c(l^*,l_g)\right]$ where the $c(a,b)$ represents the cost of reaching $b$ from $a$ while following the optimal trajectory and its value is the same used to drive the rollout process. 
%
%

Current implementation uses Python 3.6 and will be soon available online.
\section{Results}
\subsubsection{Environments}
We consider four environments: a  small and a big squared environment, a cross-like environment, a builiding like maze as shown in Fig.~\ref{fig:task-description} and big randomly generated environments of size 64x64 containing 6 rooms connected by corridors.
The small environment \textbf{[SE]}  is a $5\times5$ grid. The drone is initially positioned in the center and the responder is in one of  2 adjacent cells, but the drone does not know in which one of them. The target is positioned in one of the 4 corners. 
In total 8 equally probable initial conditions are possible, representing the initial belief state of the robot in this environment.
The large environment \textbf{[LE]}   is an $11\times11$ grid. 
Again the drone is initially positioned in the center and the responder is in one of  4 adjacent cells, but the drone is not informed of which of them. The target is positioned in one of the 16 cells in the 4  corners, 4 cell in square per corner. 
In total 64 equally probable initial conditions are possible, representing the initial belief state of the robot in this environment.
The cross-like environment \textbf{[CE]} is not convex and, thus, more demanding. It is an $11\times11$ grid with the arms of the cross 5 cells wide 
where the drone and the responder start from same location, as in condition LE. The goals are located in the corners of the ends of the cross arms. 
For the building-like environment we used a grid $15\times15$ with 4 rooms, each containing two possible target locations. 

The result for  the small environment with 1000 samples and max-depth-search=14 (average on 100 trials) are reported in Table \ref{tbl:1000Samples}.
\vskip-6ex\begin{table}%
\centering%
\caption{Performance with 1000 and 100 MC samples.}%
\footnotesize{
\begin{tabular}{|c|c|c|c|c|c|c|c|c|}%
\hline
&\multicolumn{4}{|c|}{1000 Samples}&\multicolumn{4}{|c|}{100 Samples}\\
\hline
& chES  & rrES  & thES  & dfES& chES  & rrES  & thES  & dfES\\
\hline
Success & 0.85 & 0.88 & 0.54 & \textbf{0.95}& \textbf{0.8} & 0.63 & 0.25 & 0.63  \\
\hline
Steps & 7.96 & 8.43 & 1.033 & \textbf{ 7.91} & \textbf{8.31 }& 9.12 & 11.58 & 9.74  \\
\hline
RObs & 31 & 37 & 148 & 10 & 20 & 24 & 29 & 13  \\
\hline
\end{tabular}%
\caption*{ %
\scriptsize{Average behavior over 100 trials with 1000 MC samples for each step.
\textbf{Success} is the average number of times the drone found the target before 16 steps.
In the table, \textbf{Steps }is the average number of steps necessary to the drone to find the target (or 16 if it fails).
RObs total number of times the drone met the responder before finding the target. Tests performed on the following variations of the architecture:
\textbf{rrES} reward was given also when observing the responder (0.1) other than the target (1);
\textbf{chES} reward contained an entropy bonus (-H*0.2);
\textbf{thES} reward contained an entropy bonus only during the search not during rollout (-H*0.2);
\textbf{dfES} Default, reward was given only reaching the target.}
}}%
\label{tbl:1000Samples}%
\vskip-4ex
\end{table}%
When computational resources are available the \textbf{dfES} default algorithm, not using any exploration bonus,  gave the best performance. In this case, we must note that it performs better the trivial solution of navigating through the four corners which would have an expected number of steps (2+7+11+15)/4=8.75. This is true also for the chES and rrES.
chES and rrES also observe the responder much more often than dfES; yet dfES performs better than them>. This implies that dfES makes good use of the observations of the responder (both when it finds the responder in a cell and when it doesn't).
We observe, without a statistical analysis, that the policy usually chosen by the dfES robot often consisted in (a) reaching a corner, (b) going back to the center if no target is observed, and (c) going to the next corner. This strategy may maximize the probability of encountering the responder, while  not meeting it would decrease the probability of the target to be in that direction. 
Unexpectedly, the thES version of the system presents lower performance. We attribute this to the dependence of the exploration bonus on the length of the search tree branch.

Table \ref{tbl:1000Samples} reports also the results when only MC 100 samples per step are performed.
In this case  the chES system is the most effective one, followed by rrES and dfES. chES is the least affected by the reduced number of samples. This suggest that using an entropy-based bonus may support exploration of environments when not enough time is available for construction of deep search tree. 

With the big environment and 100 samples per action the best performance are achieved by the
chES,  which, in 50\% of the trials, found the target in less than 40 steps (maximum steps allowed). thES achieved (28\%), rrES 27 \% and dfES 18\%. 

In Table \ref{tbl:100Samples-cross} the results for the building like environment are reported when simulations are run for 100 MC samples. The results consistently favour the \textbf{ehES}, while the \textbf{thES}  consistently show the worst performance.  The performance of the best \textbf{dfES} are reported and are less effective than the configurations using \textbf{ehES}. 
The approximation 
does not appear to affect the performance, while yielding a substantial speed-up in the computation. Similar results have been obtained with the randomly generated bigger environments, but could not be reported for space reasons. 
\begin{table*}%
\vskip-6ex
\centering%
\caption{Performance with 100 MC samples in building like environment.}%
\footnotesize{
\begin{tabular}{|c|c|c|c|c|c|c|c|c|}%
\hline
Approx&Rollout Pol&Rollout Act&Expl Strategy&H&steps&time&reward&reward/time \\
\hline
FALSE&snRS&bRA&ehES&gH&1959&1483&76&0.051\\
\hline
TRUE&spRS&bRA&ehES&bH&1962&258&74&0.287\\
\hline
TRUE&dpRS&sRA&ehES&bH&2029&291&74&0.254\\
\hline
TRUE&dpRS&sRA&ehES&bH&2120&307&74&0.241\\
\hline
FALSE&dpRS&sRA&ehES&gH&2058&1583&74&0.047\\
\hline
FALSE&stRS&sRA&dfES&-&2148&331&69&0.208\\
\hline
TRUE&stRS&bRA&thES&bH&2431&344&42&0.122\\
\hline
TRUE&dnRS&sRA&thES&bH&2422&298&41&0.138\\
\hline
FALSE&stRS&bRA&thES&bH&2557&2006&40&0.020\\
\hline
FALSE&dnRS&sRA&thES&bH&2650&2026&34&0.017\\
\hline
\end{tabular}}%
\caption*{%
\scriptsize{Performance over 100 trials on the cross-like environment of the first best five, the most performing system without using entropy and the worst four configurations. No data collected for the \textbf{chES} due to its computational demands. \textbf{Approx} column define if the approximated belief state \textbf{aF} (True) or the complete \textbf{cF} one are tracked. \textbf{Rollout Pol} shows the rollout policy used. \textbf{Rollout Act}  shows the Rollout action selection strategy used. \textbf{Expl Strategy} shows the Exploration Strategy employed \textbf{H} shows which entropy function was used to compute the exploration bonus, \textbf{steps} reports the number of steps in 100 trials, \textbf{time} computation time, \textbf{reward} total reward obtained and \textbf{reward/time} total reward over time, representing computational efficiency.}}%
\label{tbl:100Samples-cross}%
\vspace{-8mm}
\end{table*}%
\section{Conclusions and future work}%
Current result suggested that the idea of coupling a robot  with a more expert but constrained  human partner in the task of joint search for a target in an unknown environment may be promising. Future work will focus on optimizing the methods proposed, by integrating them with flexible SLAM algorithms, devising and testing alternative exploration bonus, employing realistic responders' models, considering additional collaborative tasks, such  as verifying that the surrounding of the responder are safe.
%
\small{
\bibliographystyle{splncs04}
\bibliography{bib_lz}
}
\end{document}